# Qualitative Measures of Ambiguity

S.K.M. Wong and Z.W. Wang
Department of Computer Science
University of Regina
Regina, Saskatchewan
Canada S4S 0A2
wong@cs.uregina.ca
zhiwei@cs.uregina.ca

## Abstract

This paper introduces a qualitative measure of ambiguity and analyses its relationship with other measures of uncertainty. Probability measures relative likelihoods, while ambiguity measures vagueness surrounding those judgments. Ambiguity is an important representation of uncertain knowledge. It deals with a different type of uncertainty modeled by subjective probability or belief.

## 1 Introduction

This paper discusses a measure of uncertainty referred to as *ambiguity*. This particular type of uncertainty has roots in the work of Keynes (1921) and Knight (1921). Knight considered two types of uncertainty, measurable uncertainty or risk which may be represented by numerical probabilities, and unmeasurable uncertainty which cannot. Although the behavioral significance of such a distinction was questioned [Arrow, 1951], Ellsberg (1961) in his seminal work argued eloquently that unmeasurable uncertainty can be better understood in terms of the notion of event ambiguity. Ambiguity is about reliability, credibility, or adequacy of information used for making choices. It is not about relative support it may give to one hypothesis as opposed to another. In other words, ambiguity cannot be expressed in terms of relative likelihoods of events.

Wallsten (1990) suggested that it is perhaps more appropriate to use the term *vagueness* to describe this type of uncertainty, because vagueness is more concerned with the specification or clarity of envisioned events than with relative likelihoods (subjective probabilities). Thus, a rare event may have high ambiguity, whereas a highly probable event can be very unambiguous. In fact, the notion of ambiguity was mentioned by Savage (1954), although it was dismissed as second-order probability. Marschak (1975) also viewed ambiguity as second-order uncertainty. However, it has been demonstrated experimentally that ambiguity is a useful concept for the understanding of choice and inference [Becker and Brownson, 1964] [Yates and Zukowski, 1976].

Recently, Fishburn (1991, 1992) considered ambiguity as a primitive concept without direct reference to likelihood, subjective probability or preference. Similar to Kolmogoroff's approach, Fishburn introduced a set of axioms to characterize a numerical measure of ambiguity, $\alpha : \mathcal{A} \to Re$, where $\mathcal{A}$ denotes a family of subsets of a finite set $\Theta$ of states with events $A, B, ... \in \mathcal{A}$. This function $\alpha$ is non-negative, vanishes at the empty set, and satisfies the complementary and submodularity axioms, namely:

$(\alpha 1)$      $\alpha(\emptyset) = 0, \quad \alpha(A) \geq 0,$

$(\alpha 2)$      $\alpha(\neg A) = \alpha(A),$

$(\alpha 3)$      $\alpha(A \cap B) + \alpha(A \cup B) \leq \alpha(A) + \alpha(B).$

Note that $(\alpha 1)$ and $(\alpha 2)$ imply $\alpha(\Theta) = 0$. Clearly, this numerical measure of ambiguity is very different from the traditional measure of uncertainty modeled by subjective probability. For example, axiom $(\alpha 2)$ is strikingly different from the Kolomogoroff axiom $\alpha(\neg A) = 1 - \alpha(A)$.

There are, of course, many advantages in using a numerical function to represent ambiguity. However, in many practical situations the required quantitative information may not be available for making decisions. In fact, in many important applications, only qualitative information is required. For example, decision rules and the inference mechanism employed in expert systems are often not expressed in the form of a numerical function. Therefore, in contrast to Fishburn's approach, we introduce in this paper a qualitative measure of ambiguity. We believe that this measure is useful for qualitative reasoning and provides a more appropriate framework for studying the algebraic structure of ambiguity.

The paper is organized as follows. In Section 2, we first define a qualitative ambiguity measure, and then introduce the notion of *interval structure* which forms the basis of our exposition. In Section 3, we present the main results which explicitly reveal the structure of ambiguity, and the close relationship between am-biguity and other measures of uncertainty.

## 2   An Ambiguity Measure

Let $\Theta$ be a set of states of the world:

$$\Theta = \{\theta_1, \theta_2, ..., \theta_m\},$$

and $2^\Theta$ denotes the family of subsets of $\Theta$:

$$2^\Theta = \{A \mid A \subseteq \Theta\}.$$

Each element $A \in 2^\Theta$ represents an *event* or a *proposition* of interest. The complement of $A$, written $\neg A$, is defined by:

$$\neg A = \{\theta \mid \theta \in \Theta, \theta \notin A\}.$$

Let $\Omega$ denote a finite set of *situations*. We define a qualitative measure of ambiguity as follows:

**Definition 1** *An ambiguity measure is a mapping,*

$$a : 2^\Theta \to 2^\Omega,$$

*satisfying:*

$(a1) \quad a(\emptyset) = \emptyset,$
$(a2) \quad a(A) = a(\neg A),$
$(a3.1) \quad a(A \cap B) \cup a(A \cup B) \subseteq a(A) \cup a(B),$
$(a3.2) \quad a(A \cap B) \cap a(A \cup B) \subseteq a(A) \cap a(B).$

Note that by (a1) and (a2), we have:

$$(a4) \quad a(\Theta) = \emptyset.$$

To motivate the above definition of ambiguity, let us assume that $A \in 2^\Theta$ is a proposition. With regard to a situation $\omega \in \Omega$, a proposition $A$ is *true* or *false*. Ideally, one would like to define a subset $i(A) \subseteq \Omega$ to indicate that $A$ is true for all $\omega \in i(A)$, and $A$ is false for all $\omega \notin i(A)$. However, due to the lack of knowledge, one may not be able to specify precisely the set of situations in which a proposition is true. Instead, one may be able to provide the *lower* and *upper* bounds of incidence for the individual propositions. In other words, one can define two mappings $\underline{f} : 2^\Theta \to 2^\Omega$ and $\bar{f} : 2^\Theta \to 2^\Omega$ to indicate the *interval* within which the truth of a proposition lies. One may interpret that the lower mapping $\underline{f}$ defines a set of *deterministic* decision rules for each proposition $A \in 2^\Theta$. That is, $A$ is *definitely true* whenever $\omega \in \underline{f}(A)$, $A$ is *definitely false* whenever $\omega \in \Omega - \bar{f}(A)$, and $A$ is *possibly true* whenever $\omega \in \bar{f}(A) - \underline{f}(A)$.

Obviously, one can define a large number of lower and upper mappings from $2^\Theta$ to $2^\Omega$. We postulate that a upper mapping $\bar{f} : 2^\Theta \to 2^\Omega$ satisfies the following axioms:

$(\bar{f}1) \quad \bar{f}(\emptyset) = \emptyset,$
$(\bar{f}2) \quad \bar{f}(\Theta) = \Omega,$
$(\bar{f}3) \quad \bar{f}(A \cup B) = \bar{f}(A) \cup \bar{f}(B),$

It can be verified that axioms $(\bar{f}1)$, $(\bar{f}2)$, and $(\bar{f}3)$ imply:

$(\bar{f}4) \quad \bar{f}(A \cap B) \subseteq \bar{f}(A) \cap \bar{f}(B).$

The corresponding lower mapping $\underline{f} : 2^\Theta \to 2^\Omega$ is defined by:

$$\underline{f}(A) = \neg \bar{f}(\neg A). \quad (1)$$

One can easily show that $\underline{f}$ satisfies the following properties:

$(\underline{f}1) \quad \underline{f}(\emptyset) = \emptyset,$
$(\underline{f}2) \quad \underline{f}(\Theta) = \Omega,$
$(\underline{f}3) \quad \underline{f}(A \cap B) = \underline{f}(A) \cap \underline{f}(B).$

and

$(\underline{f}4) \quad \underline{f}(A) \cup \underline{f}(B) \subseteq \underline{f}(A \cup B).$

We call the pair of mappings $(\underline{f}, \bar{f})$ defined above an *interval structure* [Wong et al., 1992] which was suggested as a qualitative measure of belief.

Let us now state our first observation:

**Theorem 1** *Given an interval structure $(\underline{f}, \bar{f})$, the mapping, $a : 2^\Theta \to 2^\Omega$, defined by: $\forall A \in 2^\Theta$,*

$$a(A) = \bar{f}(A) - \underline{f}(A) = \bar{f}(A) \cap \neg \underline{f}(A),$$

*is an ambiguity measure, namely, it satisfies axioms $(a1), (a2), (a3.1),$ and $(a3.2)$.*

Proof:   Axiom $(a1)$ follows directly from $(\underline{f}1)$ and $(\bar{f}1)$, i.e.,

$$a(\emptyset) = \bar{f}(\emptyset) \cap \neg \underline{f}(\emptyset) = \emptyset.$$

By the definition of upper mapping,

$$\begin{aligned}
a(\neg A) &= \bar{f}(\neg A) \cap \neg \underline{f}(\neg A) \\
&= \bar{f}(\neg A) \cap \bar{f}(A) \\
&= \neg \underline{f}(A) \cap \bar{f}(A) \\
&= a(A).
\end{aligned}$$

This means that $(a2)$ holds.

From equation (1) and axiom $(\bar{f}3)$, we obtain:

$$\begin{aligned}
&a(A) \cup a(B) \\
&= [\bar{f}(A) \cap \neg \underline{f}(A)] \cup [\bar{f}(B) \cap \neg \underline{f}(B)] \\
&= [\bar{f}(A) \cap \bar{f}(\neg A)] \cup [\bar{f}(B) \cap \bar{f}(\neg B)] \\
&= [\bar{f}(A) \cup \bar{f}(B)] \cap [\bar{f}(A) \cup \bar{f}(\neg B)] \\
&\quad \cap [\bar{f}(\neg A) \cup \bar{f}(B)] \cap [\bar{f}(\neg A) \cup \bar{f}(\neg B)] \\
&= \bar{f}(A \cup B) \cap \bar{f}(A \cup \neg B) \cap \bar{f}(\neg A \cup B) \\
&\quad \cap \bar{f}(\neg A \cup \neg B).
\end{aligned}$$

Also, it follows from $(\bar{f}3)$ and $(\bar{f}4)$:

$$\begin{aligned}
&a(A \cap B) \cup a(A \cup B) \\
&= [\bar{f}(A \cap B) \cap \neg \underline{f}(A \cap B)]
\end{aligned}$$

$$\cup [\bar{f}(A\cup B)\cap \neg \underline{f}(A\cup B)]$$
$$= [\bar{f}(A\cap B)\cap \bar{f}(\neg A\cup \neg B)]$$
$$\cup [\bar{f}(A\cup B)$$
$$\cap \bar{f}(\neg A\cap \neg B)]$$
$$= [\bar{f}(A\cap B)\cup \bar{f}(A\cup B)]$$
$$\cap [\bar{f}(A\cap B)\cup \bar{f}(\neg A\cup \neg B)]$$
$$\cap [\bar{f}(\neg A\cap \neg B)\cup \bar{f}(A\cup B)]$$
$$\cap [\bar{f}(\neg A\cap \neg B)\cup \bar{f}(\neg A\cup \neg B)]$$
$$= \bar{f}((A\cap B)\cup (A\cup B))$$
$$\cap \bar{f}((A\cap B)\cup (\neg A\cup \neg B))$$
$$\cap \bar{f}((\neg A\cap \neg B)\cup (A\cup B))$$
$$\cap \bar{f}((\neg A\cap \neg B)\cup (\neg A\cup \neg B))$$
$$= \bar{f}(A\cup B)\cap \bar{f}((A\cup \neg B)$$
$$\cap (\neg A\cup B))\cap \bar{f}(\neg A\cup \neg B)$$
$$\subseteq \bar{f}(A\cup B)\cap \bar{f}(A\cup \neg B)$$
$$\cap \bar{f}(\neg A\cup B)\cap \bar{f}(\neg A\cup \neg B)$$
$$= a(A)\cup a(B).$$

Hence, (a3.1) holds.

Note that:
$$a(A\cap B)\cap a(A\cup B)$$
$$= \bar{f}(A\cap B)\cap \neg \underline{f}(A\cap B)$$
$$\cap \bar{f}(A\cup B)\cap \neg \underline{f}(A\cup B)$$
$$\subseteq \bar{f}(A\cap B)\cap \neg \underline{f}(A\cup B)$$
$$= \bar{f}(A\cap B)\cap \underline{f}(\neg A\cap \neg B),$$

and
$$a(A)\cap a(B)$$
$$= \bar{f}(A)\cap \neg \underline{f}(A)\cap \bar{f}(B)\cap \neg \underline{f}(B)$$
$$= \bar{f}(A)\cap \bar{f}(\neg A)\cap \bar{f}(B)\cap \bar{f}(\neg B)$$
$$\supseteq \bar{f}(A\cap B)\cap \bar{f}(\neg A\cap \neg B).$$

That is, (a3.2) holds. □

## 3 Relationship between Ambiguity and Other Measures of Uncertainty

Theorem 1 clearly indicates that there exists a close connection between the notions of interval structure and ambiguity. We will explore in more detail the relationship between ambiguity and other measures of uncertainty in this section.

As mentioned earlier, if one has sufficient knowledge, one would be able to define a subset of situation $i(A)\subseteq \Omega$ such that $A$ is true for all $\omega \in i(A)$, and $A$ is false for all $\omega \notin i(A)$. In this case, there would exist no ambiguity at all in characterizing the proposition $A$. Now let us define an *incidence* mapping $i$ more precisely [Bundy, 1985].

**Definition 2** *A function $i : 2^\Theta \to 2^\Omega$ is called an incidence mapping, if it satisfies:*

$(i1)\quad i(\emptyset)=\emptyset,$
$(i2)\quad i(\Theta)=\Omega,$
$(i3)\quad i(A\cup B)=i(A)\cup i(B),$
$(i4)\quad \neg i(A)=i(\neg A).$

Note that by $(i3)$ and $(i4)$, we have:
$$i(A\cap B) = \neg i(\neg (A\cap B))$$
$$= \neg i(\neg A\cup \neg B)$$
$$= \neg [i(\neg A)\cup i(\neg B)]$$
$$= \neg i(\neg A)\cap \neg i(\neg B)$$
$$= i(A)\cap i(B).$$

That is,

$(i3')\quad i(A\cap B)=i(A)\cap i(B).$

Recall that an interval structure $(\underline{f},\bar{f})$ may be viewed as a pair of mappings which provide the lower and upper bounds of incidence for the individual propositions. In fact, when the lower mapping $\underline{f}$ is identical to the upper mapping $\bar{f}$, then $i = \underline{f} = \bar{f}$ becomes an incidence mapping. Thus, one would expect in general that there exists an incidence mapping bounded by an interval structure, i.e., $\underline{f}(A)\subseteq i(A)\subseteq \bar{f}(A), \forall A\in 2^\Theta$. (See Lemma 2 in the Appendix).

We say that an ambiguity measure $a$ is *compatible* with an incidence mapping $i$, i.e., $(i,a)$ is a compatible pair if
$$a(A)\cup a(B)\subseteq i(A\cup B)\cup a(A\cup B),\quad \forall A,B\in 2^\Theta.$$

We have shown in Theorem 1 that for a given interval structure $(\underline{f},\bar{f})$, the mapping defined by $\bar{f}(A)-\underline{f}(A), \forall A\in 2^\Theta$, is an ambiguity function. This observation and Lemma 1 strongly suggest that the notions of interval structure, incidence mapping, and ambiguity measure are inter-related. The connection between these qualitative measures and other quantitative measures of uncertainty is also explored. Our results are summarized in the following two theorems.

**Theorem 2** *A pair of mappings $(\underline{f},\bar{f})$ from $2^\Theta$ to $2^\Omega$ is an interval structure, if and only if there exists a compatible pair $(i,a)$ of incidence and ambiguity mappings such that: $\forall A\in 2^\Theta$,*

$(c1)\quad \bar{f}(A) = i(A)\cup a(A),$
$(c2)\quad \underline{f}(A) = i(A)\cap \neg a(A).$

Proof: ($\Leftarrow$) By axioms (a1) and (i1),
$$\bar{f}(\emptyset) = i(\emptyset)\cup a(\emptyset) = \emptyset \cup \emptyset = \emptyset.$$
By axioms (a2) and (i2),
$$\bar{f}(\Theta) = i(\Theta)\cup a(\Theta) = \Omega \cup \Omega = \Omega.$$
By axioms (a3.1) and (i3),
$$\bar{f}(A\cup B) = i(A\cup B)\cup a(A\cup B)$$
$$\subseteq i(A\cup B)\cup a(A)\cup a(B)$$
$$= i(A)\cup i(B)\cup a(A)\cup a(B)$$
$$= [i(A)\cup a(A)]\cup [i(B)\cup a(B)]$$
$$= \bar{f}(A)\cup \bar{f}(B).$$

From equation (2) and axioms (i3), it follows:
$$\begin{aligned}\bar{f}(A\cup B) &= i(A\cup B)\cup a(A\cup B)\\ &\supseteq i(A\cup B)\cup a(A)\cup a(B)\\ &= i(A)\cup i(B)\cup a(A)\cup a(B)\\ &= [i(A)\cup a(A)]\cup [i(B)\cup a(B)]\\ &= \bar{f}(A)\cup \bar{f}(B).\end{aligned}$$

Thus,
$$\bar{f}(A\cup B) = \bar{f}(A)\cup \bar{f}(B).$$

By (i4) and (a2),
$$\begin{aligned}\bar{f}(A) &= i(A)\cup a(A)\\ &= \neg[\neg i(A)\cap \neg a(A)]\\ &= \neg[i(\neg A)\cap a(\neg A)]\\ &= \neg \underline{f}(\neg A).\end{aligned}$$

Therefore, $(\underline{f}, \bar{f})$ is an interval structure.

($\Rightarrow$) Here, we want to show that there exist mappings $i$ and $a$ such that $\underline{f}$ and $\bar{f}$ can be expressed as:
$$\bar{f}(A) = i(A)\cup a(A),$$
$$\underline{f}(A) = i(A)\cap \neg a(A),$$
and
$$a(A)\cup a(B) \subseteq a(A\cup B)\cup i(A\cup B), \quad \forall A,B \in 2^{\Theta}.$$

By Lemma 1, there exists an incidence mapping $i$ satisfying:
$$\underline{f}(A) \subseteq i(A) \subseteq \bar{f}(A), \quad \forall A\in 2^{\Theta}.$$

Let $a(A) = \bar{f}(A)\cap \neg \underline{f}(A), \forall A\in 2^{\Theta}$. By Theorem 1, this mapping $a$ is an ambiguity measure. Thus,
$$\begin{aligned}i(A)\cup a(A) &= i(A)\cup (\bar{f}(A)\cap \neg \underline{f}(A))\\ &= (i(A)\cup \bar{f}(A))\cap (i(A)\cup \neg \underline{f}(A))\\ &= \bar{f}(A)\cap (i(A)\cup \neg \underline{f}(A)).\end{aligned}$$

Note that:
$$i(A)\supseteq \underline{f}(A) \Longrightarrow i(A)\cup \neg \underline{f}(A) \supseteq \underline{f}(A)\cup \neg \underline{f}(A) = \Omega.$$

Therefore, we obtain:
$$\bar{f}(A)\cap (i(A)\cup \neg \underline{f}(A)) = \bar{f}(A)\cap \Omega = \bar{f}(A).$$

That is,
$$\bar{f}(A) = i(A)\cup a(A).$$

Similarly, we have:
$$\begin{aligned}i(A)\cap \neg a(A) &= i(A)\cap \neg(\bar{f}(A)\cap \neg \underline{f}(A))\\ &= i(A)\cap (\neg \bar{f}(A)\cup \underline{f}(A))\\ &= (i(A)\cap \neg \bar{f}(A))\cup (i(A)\cap \underline{f}(A))\\ &= (i(A)\cap \neg \bar{f}(A))\cup \underline{f}(A)\\ &= \emptyset \cup \underline{f}(A)\\ &= \underline{f}(A).\end{aligned}$$

That is,
$$\underline{f}(A) = i(A)\cap \neg a(A).$$

Also, $\bar{f}(A) = i(A)\cup a(A) = \neg(i(A)\cap \neg a(A)) = \neg \underline{f}(A)$, and
$$\begin{aligned}a(A)\cup a(B) &= (\bar{f}(A)\cap \neg \underline{f}(A))\cup (\bar{f}(B)\cap \neg \underline{f}(B))\\ &\subseteq \bar{f}(A)\cup \bar{f}(B)\\ &= \bar{f}(A\cup B)\\ &= i(A\cup B)\cup a(A\cup B). \quad \Box\end{aligned}$$

We close the present discussion by deriving the connection between the *qualitative* measures: incidence, interval structure, and ambiguity mappings, and the *quantitative* measures: probability, belief, and ambiguity functions.

**Theorem 3** *A pair of functions Bel and Pl from $2^{\Theta}$ to $[0,1]$ are belief and plausible functions, if and only if there exist a compatible pair $(i, a)$ of incidence and ambiguity mappings from $2^{\Theta}$ to $2^{\Omega}$, and a probability function $P$ on $2^{\Omega}$ such that:*
$$Bel(A) = P(i(A)\cap \neg a(A))$$
*and*
$$Pl(A) = P(i(A)\cup a(A)).$$

Proof: ($\Leftarrow$) By Theorem 2, we can construct an interval structure $(\underline{f}, \bar{f})$ from a compatible pair $(i, a)$ of incidence and ambiguity mappings. By Lemma 1, $(\underline{f}, \bar{f})$ can be expressed in terms of a basic assignment $j$. Let $m(B) = P(j(B)), \forall B \in 2^{\Theta}$. It follows from $(j1), (j2), (j3)$, and the Kolmogoroff axioms:
$$m(\emptyset) = P(j(\emptyset)) = P(\emptyset) = 0,$$
$$\sum_{B\subseteq \Theta} m(B) = \sum_{B\subseteq \Theta} P(j(B)) = P(\bigcup_{B\subseteq \Theta} j(B)) = P(\Omega) = 1,$$
and
$$P(\underline{f}(A)) = P(\bigcup_{B\subseteq A} j(B)) = \sum_{B\subseteq \Theta}(P(j(B))) = \sum_{B\subseteq \Theta} m(B).$$

Thus, the function, $Bel: 2^{\Theta} \to [0,1]$, defined by:
$$Bel(A) = P(\underline{f}(A)), \quad \forall A \in 2^{\Theta},$$
is a belief function, and $Pl(A) = P(\bar{f}(A))$ is the corresponding plausibility function.

($\Rightarrow$) Suppose $Bel: 2^{\Theta} \to [0,1]$ is a belief function. There exists a basic probability assignment $m: 2^{\Theta} \to [0,1]$ such that $Bel(A) = \sum_{B\subseteq A} m(B)$. Each subset $B \in 2^{\Theta}$ with $m(B) \neq 0$ is called a focal element. Using the focal elements, we can construct a finite set $\Omega$ such that each $\omega \in \Omega$ corresponds to a focal element $B$. Thus, the elements in $\Omega$ can be uniquely labeled by the focal elements, namely:
$$\Omega = \{\omega_B \mid B \text{ is a focal element}\}.$$

We can define a probability function on $2^{\Omega}$ as:
$$P(\{\omega_B\}) = m(B).$$

One can construct a basic assignment mapping $j : 2^\Theta \to 2^\Omega$ as:

$$j(B) = \begin{cases} \{\omega_B\} & \text{if } m(B) \neq 0, \\ \emptyset & \text{if } m(B) = 0. \end{cases}$$

Let $\underline{f}(A) = \bigcup_{B \subseteq A} j(B)$ and $\bar{f}(A) = \neg \underline{f}(\neg A), \forall A \in 2^\Theta$. By Lemma 1, we can immediately conclude that the pair of mappings $(\underline{f}, \bar{f})$ is an interval structure. Moreover, $\forall A \in 2^\Theta$,

$$Bel(A) = P(\underline{f}(A)$$

and

$$Pl(A) = P(\bar{f}(A)).$$

The desired results immediately follow from Theorem 2. □

It should be noted that $P(i(A))$ is a probability function on $2^\Theta$. The function defined by: $\forall A \in 2^\Theta$,

$$\alpha(A) = P(a(A)) = Pl(A) - Bel(A)$$

is, in fact, an ambiguity function, as it satisfies the axioms ($\alpha 1$) to ($\alpha 3$) suggested by Fishburn [Fishburn, 1992].

## 4  Conclusion

In this paper, we have introduced a qualitative measure of ambiguity. This notion of ambiguity is similar to that considered by Ellsberg. Probability measures the relative judgments of likelihoods, whereas ambiguity measures the uncertainty or vagueness surrounding those judgments. Both probability and ambiguity are important representations of uncertain knowledge, although they deal with different types of uncertainty. We have also established the inter-relationship between ambiguity, belief, and probability measures from both the *qualitative* and *quantitative* points of view.

The results of this preliminary study suggest that it is useful to further explore the theoretical and practical issues in using ambiguity for uncertain reasoning.

## Appendix

**Lemma 1** $(\underline{f}, \bar{f})$ *is an interval structure if and only if there exists a basic assignment* $j : 2^\Theta \to 2^\Omega$ *satisfying:*

$(j1)$  $j(\emptyset) = \emptyset$,

$(j2)$  $\bigcup_{B \subseteq \Theta} j(B) = \Omega$,

$(j3)$  $A \neq B \Rightarrow j(A) \cap j(B) = \emptyset$,

*such that:*

$$\underline{f}(A) = \bigcup_{B \subseteq A} j(B), \quad \forall A \in 2^\Theta,$$

*and*

$$\bar{f}(A) = \neg \underline{f}(\neg A).$$

Proof: ($\Rightarrow$) Given an interval structure $(\underline{f}, \bar{f})$, we can construct a mapping $j : 2^\Theta \to 2^\Omega$ as follows: $\forall A \in 2^\Theta$,

$$j(A) = \underline{f}(A) - \bigcup_{B \subset A} \underline{f}(B).$$

First, let us show that:

$$\underline{f}(A) = \bigcup_{B \subseteq A} j(B).$$

Suppose $\omega \in \bigcup_{B \subseteq A} j(B)$. This means that there exists a $\Gamma \subseteq A$ such that $\omega \in j(\Gamma)$. By the definition of $j$, $\omega \in \underline{f}(\Gamma)$. Since $(\underline{f}, \bar{f})$ is an interval structure, by ($\underline{f}4$),

$$\underline{f}(A) \cup \underline{f}(\Gamma) \subseteq \underline{f}(A \cup \Gamma).$$

As $\Gamma \subseteq A$, we have:

$$\underline{f}(A) \cup \underline{f}(\Gamma) \subseteq \underline{f}(A).$$

Consequently, $\omega \in \underline{f}(A)$, i.e.,

$$\underline{f}(A) \supseteq \bigcup_{B \subseteq A} j(B).$$

Conversely, suppose $\omega \in \underline{f}(A)$. There are two possibilities: $\omega \in j(A)$ or $\omega \notin j(A)$. If $\omega \in j(A)$, then $\omega \in \bigcup_{B \subseteq A} j(B)$. If $\omega \notin j(A)$, according to the definition of $j(A)$, we have $\omega \in \bigcup_{B \subset A} \underline{f}(B)$. Thus,

$$\exists \Gamma \subset A \text{ such that } \omega \in \underline{f}(\Gamma).$$

In this case, $\omega \in j(\Gamma)$ or $\omega \notin j(\Gamma)$. Suppose $\omega \in j(\Gamma)$. Since $\Gamma$ is a subset of $A$, then $\omega \in \bigcup_{B \subseteq A} j(B)$. On the other hand, if $\omega \notin j(\Gamma)$, according to the definition of $j(\Gamma)$, we have $\omega \in \bigcup_{B \subset \Gamma} \underline{f}(B)$. Therefore,

$$\exists \Delta \subset \Gamma \subset A \text{ such that } \omega \in \underline{f}(\Delta).$$

As $2^\Theta$ is a finite set, this reduction process cannot continue forever. In other words, there must be a set $E \subset A$ such that $\omega \in j(E)$. That is, $\omega \in \bigcup_{B \subseteq A} j(B)$. Thus, we obtain:

$$\underline{f}(A) \subseteq \bigcup_{B \subseteq A} j(A).$$

Now, we prove that $j$ satisfies axioms (j1), (j2), and (j3).

Since $j(\emptyset) = \underline{f}(\emptyset) - \bigcup_{B \subset \emptyset} j(B) = \underline{f}(\emptyset) = \emptyset$, and $\bigcup_{B \subseteq \Theta} j(B) = \underline{f}(\Theta) = \Omega$, (j1) and (j2) hold.

To prove (j3), we consider two separate cases of $A \neq B$:

$(i)$  $A \not\subseteq B$ and $B \not\subseteq A$,

$(ii)$  $A \subseteq B$ or $B \subseteq A$.

In case (i),

$$\begin{aligned} j(A) &= [\underline{f}(A) - \bigcup_{\Gamma \subset A} \underline{f}(\Gamma)] \\ &\subseteq \underline{f}(A) - \underline{f}(A \cap B) \\ &= \underline{f}(A) \cap \neg \underline{f}(A \cap B) \end{aligned}$$

and
$$j(B) = [\underline{f}(B) - \bigcup_{\Delta \subset B} \underline{f}(\Delta)]$$
$$\subseteq \underline{f}(B) - \underline{f}(B \cap A) = \underline{f}(B) \cap \neg \underline{f}(B \cap A).$$

Also,
$$[\underline{f}(A) \cap \neg \underline{f}(A \cap B)] \cap [\underline{f}(B) \cap \neg \underline{f}(B \cap A)]$$
$$= \underline{f}(A) \cap \underline{f}(B) \cap \neg \underline{f}(A \cap B)$$
$$= \underline{f}(A \cap B) \cap \neg \underline{f}(A \cap B)$$
$$= \emptyset.$$

Thus,
$$j(A) \cap j(B) = \emptyset.$$

In case (ii), if $A \subset B$, we obtain:
$$\underline{f}(A) \cap j(B)$$
$$= \underline{f}(A) \cap [\underline{f}(B) - \bigcup_{\Gamma \subset B, \Gamma \neq A} \underline{f}(\Gamma)]$$
$$= \underline{f}(A) \cap [\underline{f}(B) - (\underline{f}(A) \cup (\bigcup_{\Gamma \subset B, \Gamma \neq A} \underline{f}(\Gamma))]$$
$$= \underline{f}(A) \cap [\underline{f}(B) \cap \neg(\underline{f}(A) \cup (\bigcup_{\Gamma \subset B, \Gamma \neq A} \underline{f}(\Gamma))]$$
$$= \underline{f}(A) \cap [\underline{f}(B) \cap \neg \underline{f}(A) \cap \neg(\bigcup_{\Gamma \subset B, \Gamma \neq A} \underline{f}(\Gamma))]$$
$$= \underline{f}(A) \cap \neg \underline{f}(A) \cap [\underline{f}(B) \cap \neg(\bigcup_{\Gamma \subset B, \Gamma \neq A} \underline{f}(\Gamma))]$$
$$= \emptyset \cap [\underline{f}(B) \cap \neg(\bigcup_{\Gamma \subset B, \Gamma \neq A} \underline{f}(\Gamma))]$$
$$= \emptyset.$$

Since $j(A) \subseteq \underline{f}(A)$, it immediately follows:
$$j(A) \cap j(B) = \emptyset.$$

Similarly, if $B \subset A$, one can show that $j(A) \cap j(B) = \emptyset$. We can therefore conclude that (j3) holds.

($\Leftarrow$) We want to show that $(\underline{f}, \bar{f})$ is an interval structure.

By $(j1)$ and $(j2)$, we have:
$$\underline{f}(\emptyset) = \bigcup_{B \subseteq \emptyset} j(B) = j(\emptyset) = \emptyset$$

and
$$\underline{f}(\Theta) = \bigcup_{B \subseteq \Theta} j(B) = \Omega.$$

That is, axioms (f1) and (f2) are satisfied.

Note that:
$$\underline{f}(A) \cap \underline{f}(B)$$
$$= [\bigcup_{\Gamma \subseteq A} j(\Gamma)] \cap [\bigcup_{\Delta \subseteq B} j(\Delta)]$$
$$= \bigcup_{\Gamma \subseteq A, \Delta \subseteq B} j(\Gamma) \cap j(\Delta).$$

According to (j3), i.e., $j(\Gamma) \cap j(\Delta) = \emptyset$, if $\Gamma \neq \Delta$, we obtain:
$$\bigcup_{\Gamma \subseteq A, \Delta \subseteq B} j(\Gamma) \cap j(\Delta) = \bigcup_{\Gamma \subseteq A \cap B} j(\Gamma) \cap j(\Gamma)$$
$$= \bigcup_{\Gamma \subseteq A \cap B} j(\Gamma)$$
$$= \underline{f}(A \cap B).$$

Therefore, $(\underline{f}3)$ holds.

By construction, $\bar{f}(A) = \neg \underline{f}(A), \forall A \in 2^\Theta$. Thus, the pair $(\underline{f}, \bar{f})$ is an interval structure. □

**Lemma 2** *For any interval structure $(\underline{f}, \bar{f})$, there exists an incidence mapping, $i : 2^\Theta \to 2^\Omega$, such that:*
$$\underline{f}(A) \subseteq i(A) \subseteq \bar{f}(A), \qquad \forall A \in 2^\Theta.$$

Proof:  Before proving this lemma, we first introduce some new concepts. Let $C$ denote a relation between $\Theta$ and $\Omega$. If an element $\theta \in \Theta$ is related to an element $\omega$ in $\Omega$, we write $\theta C \omega$.

Let $\omega$ be an element in $\Omega$, the *preimage* $[\omega^{-1}]_C$ of $\omega$ under relation $C$ is a subset of $\Theta$ defined by:
$$[\omega^{-1}]_C = \{\theta \in \Theta \mid \theta C \omega\}.$$

Given an arbitrary relation $C$ between $\Theta$ and $\Omega$, we can always construct two mappings as follows: $\forall A \in 2^\Theta$,
$$\underline{C}(A) = \{\omega \mid [\omega^{-1}]_C \subseteq A\}$$
and
$$\bar{C}(A) = \{\omega \mid [\omega^{-1}]_C \cap A \neq \emptyset\}.$$

Since $(\underline{f}, \bar{f})$ is an interval structure, by Lemma 1, we can construct a basic assignment $j : 2^\Theta \to 2^\Omega$ such that $\forall A \in 2^\Theta$
$$\underline{f}(A) = \bigcup_{B \subseteq A} j(B)$$
and
$$\bar{f}(A) = \bigcup_{B \cap A \neq \emptyset} j(B).$$

Using the above $j$, we can construct a relation $C$ between $\Theta$ and $\Omega$ as:
$$\theta C \omega \iff \theta \in B \text{ and } \omega \in j(B).$$

This implies that
$$[\omega^{-1}]_C = B \iff \omega \in j(B).$$

Thus,
$$\underline{C}(A) = \{\omega \mid [\omega^{-1}]_c \subseteq A\}$$

$$\begin{aligned}
&= \bigcup_{B=[\omega^{-1}]_C \subseteq A} \{\omega\} \\
&= \bigcup_{B \subseteq A} (\cup_{[\omega^{-1}]_C = B} \{\omega\}) \\
&= \bigcup_{B \subseteq A} (\cup_{\omega \in j(B)} \{\omega\}) \\
&= \bigcup_{B \subseteq A} j(B) \\
&= \underline{f}(A).
\end{aligned}$$

Similarly, one can show that $\bar{C}(A) = \bar{f}(A)$, $\forall A \in 2^\Theta$.

Using the relation $C$, we can construct another relation $C'$ between $\Theta$ and $\Omega$ as follows: for every $\omega \in \Omega$, if $\omega \in j(B)$, select only one $\theta$ in $B$ such that $\theta C' \omega$.

The following observations are in order:

(i) In $C'$, every $\omega \in \Omega$ is related to one and only one $\theta \in \Theta$. In other words, $C'$ is a function from $\Omega$ to $\Theta$. However, two different $\omega$'s may be mapped to the same $\theta$.

(ii) Note that $[\omega^{-1}]_{C'} = \{\theta \mid \theta C' \omega\}$ is a singleton set. This means that $\forall A \in 2^\Theta$,

$$\underline{C}'(A) = \bigcup_{[\omega^{-1}]_{C'} \subseteq A} \{\omega\} = \bigcup_{[\omega^{-1}]_{C'} \cap A \neq \emptyset} \{\omega\} = \bar{C}'(A).$$

(iii)
$$\underline{C}'(A) = \bigcup_{\{\theta\}=[\omega^{-1}]_{C'} \subseteq A} \{\omega\} = \bigcup_{\{\theta\} \subseteq A} j'(\{\theta\}),$$

where
$$j'(\{\theta\}) = \{\omega \mid [\omega]_{C'} = \{\theta\}\}.$$

(iv)
$$\begin{aligned}
\omega \in \underline{C}(A) &= \bigcup_{B \subseteq A} j(B) \\
&= \omega \in j(B), \; \exists B \subseteq A \\
&= \theta C' \omega, \; \exists \theta \in B \subseteq A \\
&= \{\theta\} = [\omega^{-1}]_{C'} \\
&= \omega \in j'(\{\theta\}), \; \exists \theta \in A \\
&= \omega \in \underline{C}'(A) \\
&= \bigcup_{\{\theta\} \subseteq A} j'(\{\theta\}).
\end{aligned}$$

That is,
$$\underline{C}(A) \subseteq \underline{C}'(A), \; \forall A \in 2^\Theta.$$

(v)
$$\begin{aligned}
\omega \in \bar{C}(A) &= \bigcup_{\{\theta\} \subseteq A} j'(\{\theta\}) \\
&= \omega \in j'(\{\theta\}), \; \exists \{\theta\} \subseteq A \\
&= \theta \in B, \; B \cap A \neq \emptyset \\
&= \omega \in j(B), \; B \cap A \neq \emptyset
\end{aligned}$$

$$\begin{aligned}
&= \omega \in \bar{C}(A) \\
&= \bigcup_{B \cap A \neq \emptyset} j(B),
\end{aligned}$$

namely:
$$\bar{C}'(A) \subseteq \bar{C}(A), \; \forall A \in 2^\Theta.$$

According to (iv) and (v), $\underline{C}'(A) = \bar{C}'(A)$. Let $i(A) = \underline{C}'(A) = \bar{C}'(A)$. We finally obtain:
$$\underline{f}(A) \subseteq i(A) \subseteq \bar{f}(A), \; \forall A \in 2^\Theta.$$

We now show that $i$ satisfies axioms (i1), (i2), (i3), and (i4). Note that:

$$\begin{aligned}
i(\emptyset) &= \underline{C}'(\emptyset) \\
&= \bigcup_{[\omega^{-1}]_{C'} \subseteq \emptyset} \{\omega\} \\
&= \emptyset.
\end{aligned}$$

$$\begin{aligned}
i(\Theta) &= \underline{C}'(\Theta) \\
&= \bigcup_{[\omega^{-1}]_{C'} \subseteq \Theta} \{\omega\} \\
&= \bigcup_{\omega \in \Omega} \{\omega\} \\
&= \Omega,
\end{aligned}$$

and
$$\begin{aligned}
i(A \cup B) &= \bigcup_{[\omega^{-1}]_{C'} \subseteq A \cup B} \{\omega\} \\
&= (\bigcup_{[\omega^{-1}]_{C'} \subseteq A} \{\omega\}) \cup (\bigcup_{[\omega^{-1}]_{C'} \subseteq B} \{\omega\}) \\
&= i(A) \cup i(B).
\end{aligned}$$

Thus, $(i1)$, $(i2)$, and $(i3)$ hold.

Also,
$$\begin{aligned}
\Omega &= i(\Theta) \\
&= \bigcup_{\{\theta\} \subseteq \Theta} j'(\{\theta\}) \\
&= \bigcup_{\{\theta\} \subseteq A \cup \neg A} j'(\{\theta\}) \\
&= (\bigcup_{\{\theta\} \subseteq A} j'(\{\theta\})) \cup (\bigcup_{\{\theta\} \subseteq \neg A} j'(\{\theta\}))
\end{aligned}$$

and
$$(\bigcup_{\{\theta\} \subseteq A} j'(\{\theta\})) \cap (\bigcup_{\{\theta\} \subseteq \neg A} j'(\{\theta\})) \neq \emptyset.$$

Hence,
$$\begin{aligned}
i(\neg A) &= \bigcup_{\{\theta\} \subseteq \neg A} j'(\{\theta\}) \\
&= \Omega - \bigcup_{\{\theta\} \subseteq A} j'(\{\theta\}) \\
&= \Omega - i(A) \\
&= \neg i(A).
\end{aligned}$$

That is, (i4) holds.   □